\pdfoutput=1

\documentclass[11pt]{article}

\usepackage[final]{acl}

\usepackage{times}
\usepackage{latexsym}
\usepackage{amsmath}
\usepackage{amsfonts}
\usepackage[T1]{fontenc}

\usepackage[utf8]{inputenc}

\usepackage{microtype}

\usepackage[ruled,vlined]{algorithm2e}
\SetAlFnt{\small} 
\SetAlCapFnt{\small}
\SetAlCapNameFnt{\small}
\SetAlgoNlRelativeSize{-1} 
\usepackage{booktabs}
\usepackage{tabularx}
\usepackage{algorithm2e}
\usepackage{float}
\usepackage{placeins}

\usepackage{inconsolata}
  
\usepackage{graphicx}

\title{Mechanistic Unveiling of Transformer Circuits: \\ Self-Influence as a Key to Model Reasoning}

\author{
  Lin Zhang$^{1,2,3}$\thanks{Work done during an internship at PRADA Lab.}, 
  Lijie Hu$^{1,2}$,
  Di Wang$^{1,2}$\thanks{Corresponding author.} \\
  $^1$Provable Responsible AI and Data Analytics (PRADA) Lab \\
  $^2$King Abdullah University of Science and Technology \\
  $^3$Harbin Institute of Technology, Shenzhen \\
  \texttt{23s058005@stu.hit.edu.cn} \\
  \texttt{\{lijie.hu, di.wang\}@kaust.edu.sa} \\
}

\begin{document}
\maketitle

\begin{abstract}
Transformer-based language models have achieved significant success; however, their internal mechanisms remain largely opaque due to the complexity of non-linear interactions and high-dimensional operations. While previous studies have demonstrated that these models implicitly embed reasoning trees, humans typically employ various distinct logical reasoning mechanisms to complete the same task. It is still unclear which multi-step reasoning mechanisms are used by language models to solve such tasks. In this paper, we aim to address this question by investigating the mechanistic interpretability of language models, particularly in the context of multi-step reasoning tasks. Specifically, we employ circuit analysis and self-influence functions to evaluate the changing importance of each token throughout the reasoning process, allowing us to map the reasoning paths adopted by the model. We apply this methodology to the GPT-2 model on a prediction task (IOI) and demonstrate that the underlying circuits reveal a human-interpretable reasoning process used by the model. 
\end{abstract}

\section{Introduction}

\begin{figure*}[t]
    \centering
\includegraphics[width=0.9\linewidth]{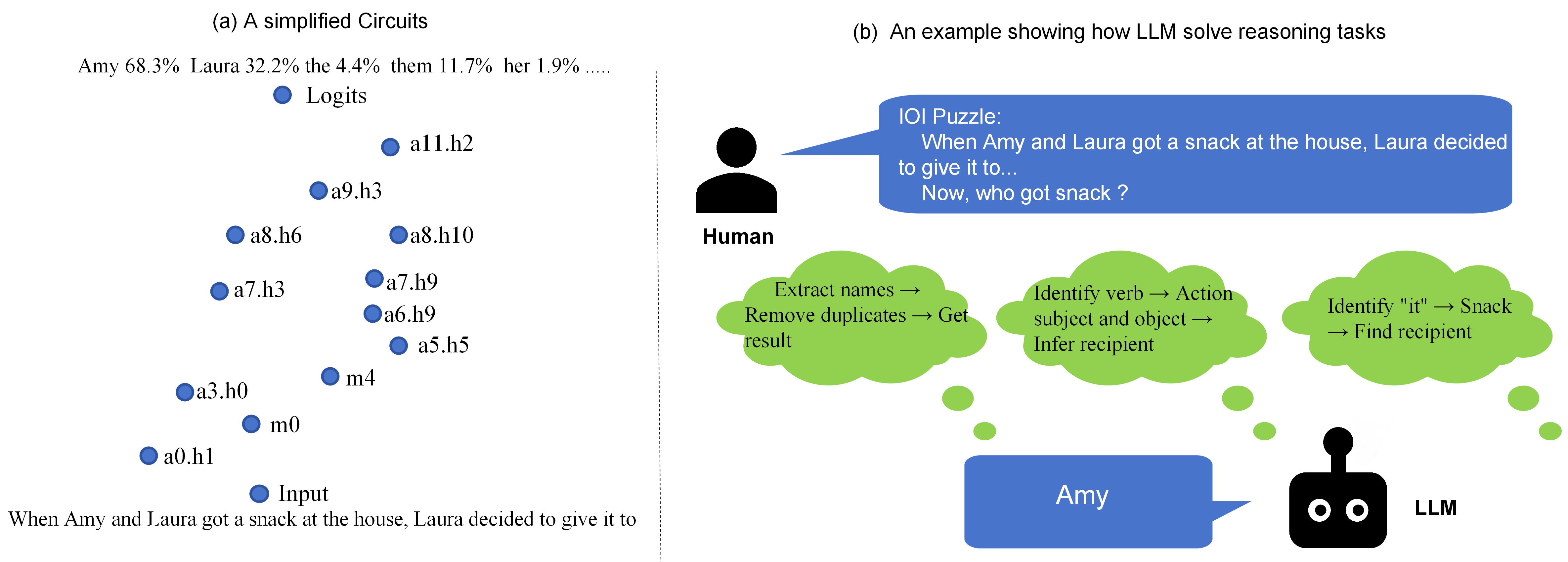}  
    \caption{(a) A simplified illustration of circuits within the model. (b) An example of how a language model (LLM) tackles a reasoning task, such as the Indirect Object Identification (IOI) puzzle. The model identifies key entities, actions, and pronouns to deduce the recipient. The reasoning process involves steps like extracting names, determining actions, and linking pronouns to objects to reach the answer "Amy."}    \label{fig:fig1}
\vspace{-12pt}
\end{figure*}

In recent years, the Transformer architecture, introduced by \citep{vaswani2017attention}, has become an efficient neural network structure for sequence modeling \citep{NEURIPS2020_1457c0d6}. Previous research \citep{hou2023towards,dong2021fly} has confirmed that large models rely primarily on reasoning rather than mere memorization when answering questions. However, the "thought process" of these models remains unclear, as shown in Figure \ref{fig:fig1}. How can we explore the thought process employed by large models during reasoning \citep{wei2022chain,kojima2022large}? Addressing this question is not only crucial for deepening our understanding of these models but also essential for developing the next generation of reliable language-based reasoning systems \citep{creswell2022faithful,creswell2022selection,chen2023reckoning,hu2024hopfieldian,cheng2024multi,yang2024moral}.

Influence functions focus on analyzing models from the perspective of their training data \citep{koh2017understanding}. They have been shown to be versatile tools applicable to a wide range of tasks, including understanding model behavior, debugging models, detecting dataset errors, and generating visually indistinguishable adversarial examples\cite{hu2024editable,hu2024dissecting}. As a variant of influence functions, self-influence is a technique for evaluating the impact of specific inputs within a neural network on the model's output. For different tokens within an input sample, self-influence scores reflect the significance of each token across various layers of the model. By tracking the influence changes of different tokens throughout the reasoning process of large language models (LLMs), it is possible to map the thought process executed by the model. However, directly calculating self-influence for all parameters in LLMs is practically infeasible due to the enormous computational resources and substantial memory consumption required.

As a result, circuit analysis within the framework of Mechanistic Interpretability (MI) \citep{olah2022mechanistic,nanda2023mechanistic} has become a focal point of research. MI aims to discover, understand, and verify the algorithms encoded in model weights by reverse engineering the model's computations into human-understandable components \citep{meng2022locating,geiger2021causal,geva2020transformer,zhang2024locate,hong2024dissecting}. A key method in this field is circuit analysis \citep{conmy2023towards,olah2020zoom}. In this approach, neural networks are conceptualized as computational graphs, where circuits represent sub-graphs composed of interconnected features and the weights that link them. These circuits function as fundamental computational units and building blocks of the network \citep{michaud2024quantization,olah2020zoom}. Representing the most critical components for completing specific tasks, these circuits capture essential computational processes without the complexity of analyzing the full model.

To address the aforementioned limitations, we propose the Self-Influence Circuit Analysis Framework (SICAF) as a mechanistic interpretation method to trace and analyze the thought process that language models (LMs) employ during complex reasoning tasks. To facilitate detailed analysis, our investigation proceeds in three stages: (1) identifying the circuit within the model through existing automatic circuit-finding methods, (2) calculating the self-influence of different tokens across various layers of the circuit in each sample, and (3) deducing the thought process that the language model employs during reasoning by analyzing changes in self-influence scores across layers.

As a preliminary step, we employ the automatic circuit-finding methods EAP, EAP-IG, and EAP-IG-KL \citep{hanna2024have} with varied parameters to identify circuits within a (finetuned) GPT-2 model \citep{radford2019language} that performs a specific natural language task, indirect object identification (IOI) \citep{wang2022interpretability}. We find that these circuits are small (containing 1-2\% of edges) and faithful (recovering $\geq$85\% of model performance). Under the same parameter constraints, EAP-IG identifies circuits that are more faithful. Furthermore, the nodes within these circuits are primarily concentrated in the first layer and the last few layers of the model.

Next, we conduct a detailed examination of the nodes within the identified circuits, calculating the self-influence of different tokens across various layers of each sample and inferring the reasoning process the language model employs by analyzing how self-influence scores change across layers.

By focusing on a specific task within the (finetuned) GPT-2 model, we have gained several key insights into the challenges of mechanistic interpretability in transformer-based language models. In particular:

\begin{itemize}
\item We propose a new mechanistic interpretation framework, SICAF, to trace and analyze the thought process that language models (LMs) employ during complex reasoning tasks. 
\item By employing various methods to identify and analyze circuits within the model, we observed a consistent pattern: the model’s key parameters are primarily concentrated in the first layer and the final few layers. 
\item We extend SICAF by applying multiple Circuit Analysis methods to uncover and analyze diverse thought processes embedded in different circuits of the model.
\end{itemize}

\section{Related Work}
\paragraph{Interpretability Methods in Language Models.} Interpretability paradigms for AI decision-making range from black-box techniques, which focus on input-output relationships, to internal analyses that delve into model mechanics \citep{bereska2024mechanistic}. Behavioral interpretability \citep{warstadt2020blimp,covert2021explaining,casalicchio2018visualizing} treats models as black boxes, examining robustness and variable dependencies, while attributional interpretability \citep{sundararajan2017axiomatic,smilkov2017smoothgrad,shrikumar2017learning} traces outputs back to individual input contributions. Concept-based interpretability \citep{belinkov2022probing,burns2023discovering,zou2023representation,yang2024makes,huimproving} explores high-level concepts within models' learned representations. In contrast, mechanistic interpretability \citep{bereska2024mechanistic} adopts a bottom-up approach, analyzing neurons, layers, and circuits to uncover causal relationships and precise computations, offering a detailed understanding of the model's internal operations.

\paragraph{Circuit Analysis.} Neural networks can be conceptualized as computational graphs, where circuits of linked features and weights serve as fundamental computational units \citep{bereska2024mechanistic}. Recent research has focused on dissecting models into interpretable circuits. Automated Circuit Discovery (ACDC) \citep{conmy2023towards} automates a large portion of the mechanistic interpretability workflow, but it is inefficient due to its recursive nature. \citet{syed2023attribution} introduced Edge Attribution Patching (EAP) to identify circuits for specific tasks, while \citet{hanna2024have} introduced EAP with integrated gradients(EAP-IG), which improves upon EAP by identifying more faithful circuits. Circuit analysis leverages key task-relevant parameters \citep{bereska2024mechanistic} and feature connections \citep{he2024dictionary} within the network to capture core computational processes and attribute outputs to specific components \citep{miller2024transformer}, bypassing the need to analyze the entire model. This approach maintains efficiency and scalability, offering a practical alternative for understanding model behavior.

\paragraph{Influence Function.} The influence function, initially a staple in robust statistics \citep{cook2000detection,cook1980characterizations}, has seen extensive adoption within machine learning since \citet{koh2017understanding} introduced it to the field. Its versatility spans various applications, including detecting mislabeled data, interpreting models, addressing model bias, and facilitating machine unlearning tasks. Notable works in machine unlearning encompass unlearning features and labels \citep{warnecke2021machine}, minimax unlearning \citep{liu2024certified}, forgetting a subset of image data for training deep neural networks \citep{golatkar2020eternal,golatkar2021mixed}, graph unlearning involving nodes, edges, and features. Recent advancements, such as the LiSSA method \citep{agarwal2017second,kwon2023datainf} and kNN-based techniques \citep{guo2021fastif}, have been proposed to enhance computational efficiency. Besides, various studies have applied influence functions to interpret models across different domains, including natural language processing \citep{han2020explaining} and image classification \citep{basu2021influence}, while also addressing biases in classification models \citep{wang2019repairing}, word embeddings \citep{brunet2019understanding}, and finetuned models \citep{chen2020multi}. Despite numerous studies on influence functions, we are the first to apply them to explain the thought process in language models (LMs) during reasoning tasks. We propose a new mechanistic interpretation framework, SICAF, to trace and analyze the reasoning strategies that language models (LMs) employ for complex tasks. Furthermore, compared to traditional neural networks, circuits contain only the most essential parameters of the model, significantly reducing the computational cost of calculating influence functions.

\section{Preliminary}
\textbf{Automate Circuit Finding.}
Edge Attribution Patching (EAP) is a gradient-based method designed to efficiently identify circuits responsible for specific behaviors in neural networks. It estimates the importance of each edge by calculating the change in the model's loss when that edge is corrupted. The score for an edge \( (u, v) \) is given by:
\begin{equation}
(z'_u - z_u)^\top \nabla_v L(s)
\end{equation}
where \( \nabla_v L(s) \) is the gradient of the loss function \( L \) with respect to the input of node \( v \), and \( z_u \) and \( z'_u \) represent the clean and corrupted inputs to node \( u \), respectively. EAP-IG extends this approach by incorporating Integrated Gradients, which computes gradients along a linear path between clean and corrupted inputs. The integrated gradients score for an edge \( (u, v) \) is:
\begin{equation}
(z'_u - z_u) \frac{1}{m} \sum_{k=1}^{m} \frac{\partial L(z' + \frac{k}{m}(z - z'))}{\partial z_v}
\end{equation}
where \( m \) is the number of steps used to approximate the integral, and \( z \) and \( z' \) represent the clean and corrupted inputs. EAP-IG addresses the issue of near-zero gradients in important features, providing a more accurate estimation of edge importance and a more faithful representation of the model's behavior. Both methods aim to identify the most crucial edges in a model's circuit, but EAP-IG achieves this with greater precision and reliability. EAP-IG-KL further runs EAP-IG with Kullback-Leibler (KL) divergence as the loss improves upon EAP-IG by incorporating Kullback-Leibler (KL) divergence to measure the difference between the model's activations on clean and corrupted inputs, ensuring higher fidelity in capturing task-specific behaviors, even under interventions.  run EAP-IG with KL divergence as the loss.

\paragraph{Influence Functions.}
Influence functions provide an efficient approximation for measuring how small perturbations in the training data affect a model’s parameters without retraining. For a model with parameters \(\theta^*\), minimizing the empirical risk over training data, the influence function evaluates how a slight change in a specific training point \(z\) modifies the model’s parameters \(\theta_{\epsilon}\) when its weight is increased by \(\epsilon\). The optimization problem is defined as:
\begin{equation}
\theta_{\epsilon}\{z\} = \arg\min_{\theta} \frac{1}{n} \sum_{i=1}^{n} \ell(h_{\theta}(z_i)) + \epsilon \ell(h_{\theta}(z))
\end{equation}
Using a first-order Taylor expansion around \(\theta^*\), the new parameters \(\theta_{\epsilon}\) can be approximated as:
\begin{equation}
\theta_{\epsilon} \approx \theta^* - \epsilon H_{\theta^*}^{-1} \nabla_{\theta} \ell(h_{\theta^*}(z))
\end{equation}
Where \(H_{\theta^*}\) is the Hessian matrix of the loss function with respect to the model parameters, and \(\nabla_{\theta} \ell(h_{\theta^*}(z))\) is the gradient of the loss function evaluated at \(\theta^*\). The influence function is then given by:
\begin{equation}
I(z) = - H_{\theta^*}^{-1} \nabla_{\theta} \ell(h_{\theta^*}(z))
\end{equation}
Additionally, the influence of a training sample \(z\) on a test sample \(z_t\) is:
\begin{equation}
I(z, z_t) = - \nabla_{\theta} \ell(h_{\theta^*}(z_t))^\top H_{\theta^*}^{-1} \nabla_{\theta} \ell(h_{\theta^*}(z))
\end{equation}
This equation measures the approximate change in the test sample’s loss when the weight of a training sample is perturbed, offering insight into how training points influence model predictions.

\section{Method}
We propose a novel mechanistic interpretation framework, SICAF, to trace and analyze the thought process that language models (LMs) employ during complex reasoning tasks. Our method comprises two primary steps: first, we apply an automatic circuit-finding approach to identify the critical circuits within the model; second, we calculate the self-influence \( I_H(x, x) \) for each layer of the circuit to assess the contribution of individual tokens to the model's decision-making process. By examining contributions at each layer, we can infer the thought process manifested by the model during reasoning. This layer-wise self-influence analysis provides a detailed understanding of the internal reasoning process while ensuring computational feasibility by focusing exclusively on the model's most essential components.

\subsection{Automatic Circuit-Finding}

To effectively locate the most important subgraphs, or circuits, in the model, we utilize advanced automatic circuit-finding methods, including Edge Attribution Patching (EAP), EAP-IG, and EAP-IG-KL. These methods allow us to identify and isolate the key circuits that contain essential features and their connecting weights, which are necessary for task completion. By focusing on these circuits rather than analyzing the entire model, we streamline the analysis and capture the primary information flow, thus attributing the model's output to specific components. This approach not only enhances efficiency but also enables us to focus on the most impactful areas of the model, laying a strong foundation for subsequent self-influence analysis.

In our approach, EAP identifies important edges by measuring the change in the loss function when each edge is perturbed, effectively constructing a "map" of the critical connections within the network. EAP-IG and EAP-IG-KL extend this by incorporating Integrated Gradients (IG) and Kullback-Leibler (KL) divergence, respectively. EAP-IG improves fidelity by more accurately capturing the importance of edges with low gradients, while EAP-IG-KL leverages KL divergence to ensure that the circuits faithfully represent the model’s behavior under various interventions. Together, these methods allow us to efficiently locate the most faithful circuits in the network, ensuring that only the most relevant parts of the model are included in the analysis.

\subsection{Self-Influence Calculation}
Once critical circuit components have been isolated, the key remaining step is to interpret the computations performed by these components. Few methods have been proposed to interpret extracted circuits. Kevin Wang et al. \citep{wang2022interpretability} explored this by knocking out a single node—a (head, token position) pair in the circuit—revealing heads with different functions. Arthur Conmy et al. \citep{conmy2023towards} proposed testing hypotheses about the functions implemented by each node in the circuit. Yunzhi Yao et al. \citep{yao2024knowledge} evaluated the impact of current knowledge editing techniques on these knowledge circuits, providing deeper insights into the functionality and limitations of these editing methodologies. We differ from these works by calculating the self-influence of each token in the sample across different layers of the circuit to reverse-engineer, which human-understandable reasoning patterns the model employs. We compute the self-influence \( I_H(x, x) \) for each layer within the circuit. The self-influence formula is defined as:
\begin{equation}
I_H(x, x) = - \nabla_{\theta} L(x)^\top H^{-1} \nabla_{\theta} L(x)
\end{equation}
where \( \nabla_{\theta} L(x) \) is the gradient of the loss function \( L(x) \) with respect to the parameters \( \theta \) in the circuit at each layer, and \( x \) represents the tokens in the sample. The Hessian matrix \( H \) represents the second-order derivatives of the loss function with respect to the parameters, and \( H^{-1} \) is its inverse. Calculating self-influence allows us to measure each token’s impact on the parameter updates, revealing the degree to which each token contributes to the decision-making process at each layer.

However, calculating the inverse of the Hessian \( H^{-1} \) directly is computationally expensive, particularly in large-scale models. To mitigate this, we adopt a divide-and-conquer strategy by utilizing the Hessian-vector product (HVP) in place of the explicit inverse calculation. The HVP approach allows us to approximate \( H^{-1}v \) without direct inversion by first calculating \( \frac{\partial f}{\partial x}(x) \), where \( x \) represents tokens in the sample, and then computing:
\begin{equation}
\frac{\partial x}{\partial} \left( \frac{\partial f(x)}{\partial x} \cdot v \right)
\end{equation}
where \( \frac{\partial f}{\partial x}(x) \in  \mathbb{R}^{1 \times d} \) represents the gradient of the function \( f(x) \) with respect to the input tokens, and \( v \in \mathbb{R}^{d \times 1} \) is a vector. This product \( \left( \frac{\partial f(x)}{\partial x} \cdot v \right) \) is a scalar, and computing its gradient with respect to \( x \) is computationally efficient in deep learning frameworks like PyTorch and TensorFlow.

To further approximate \( H^{-1}v \), we leverage a recursive Taylor expansion:
\begin{equation}
H^{-1} = \sum_{i=0}^{\infty} (I - H)^i
\end{equation}
This expansion enables us to iteratively compute $H^{-1}v$, avoiding the computational expense of explicit inversion. Additionally, to ensure $ ||H|| \leq 1 $, we scale the Hessian $H$ by a factor $c \in \mathbb{R}_{+}$, allowing us to approximate $H^{-1}$ as $ c(cH)^{-1}$, which further reduces computational complexity. This approach enables efficient estimation of self-influence values, preserving the practicality of layer-wise influence analysis for large models.

With the self-influence values calculated across all layers within the circuit, we can now trace the flow of information and identify how different tokens contribute to the model’s decision-making at each layer. By examining these contributions in a layer-by-layer fashion, we are able to infer the structure of the "reasoning tree" that the model implicitly follows during inference. This reasoning tree structure elucidates the hierarchical process by which the model accumulates and combines information, offering insights into the specific patterns of reasoning the model employs.

Our layer-wise self-influence analysis provides a comprehensive view of the internal mechanisms that drive the model's behavior. By focusing on critical components within the network, our method maintains computational feasibility while offering a fine-grained understanding of the model’s decision-making process. This approach not only unveils the underlying reasoning patterns but also provides a valuable theoretical foundation for improving and optimizing transformer-based models for reasoning tasks. 

As shown in Algorithm~\ref{alg:alg1}, this process of Circuit-Based Self-Influence Analysis allows us to construct the reasoning tree structure effectively.

\section{Experiment}
\subsection{Experimental Setting}

\begin{figure*}[ht]
    \centering
\includegraphics[width=0.8\linewidth]{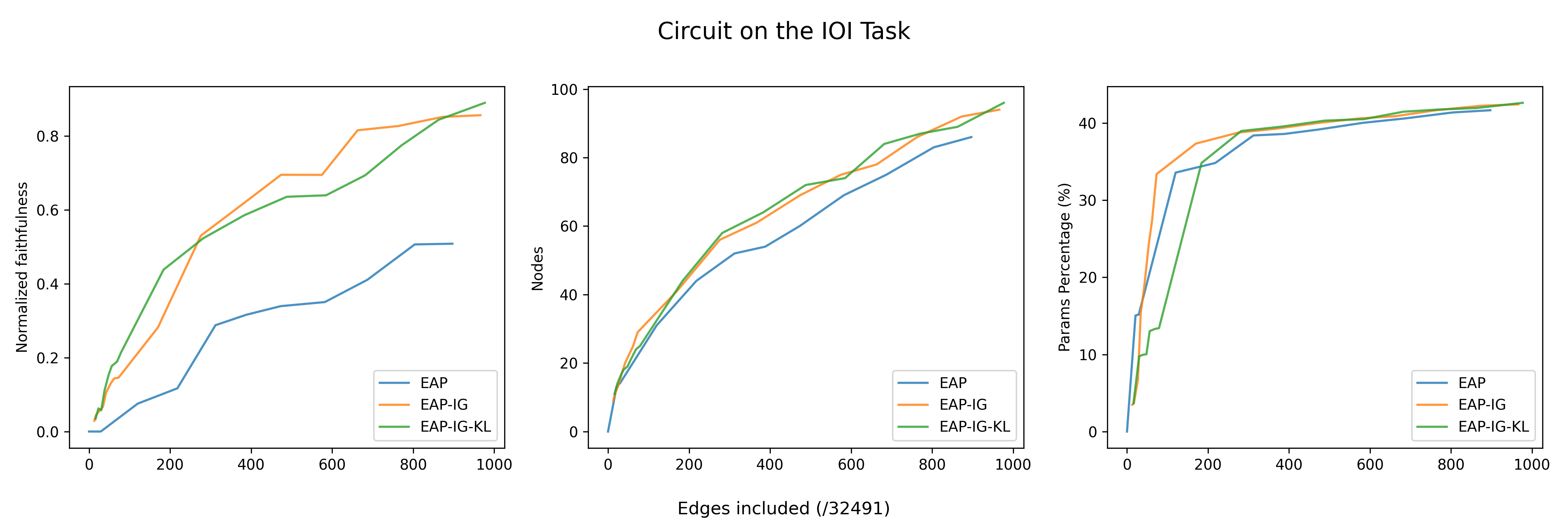}  
    \caption{Comparison of normalized faithfulness, number of nodes, and parameter percentage for circuits identified by EAP, EAP-IG, and EAP-IG-KL on the IOI task. The x-axis represents the number of edges included, and each panel shows different metrics: normalized faithfulness (left), number of nodes (middle), and parameter percentage (right).}
    \label{fig:fig2}
\vspace{-12pt}
\end{figure*}

\begin{figure*}[ht]
    \centering
\includegraphics[width=0.8\linewidth]{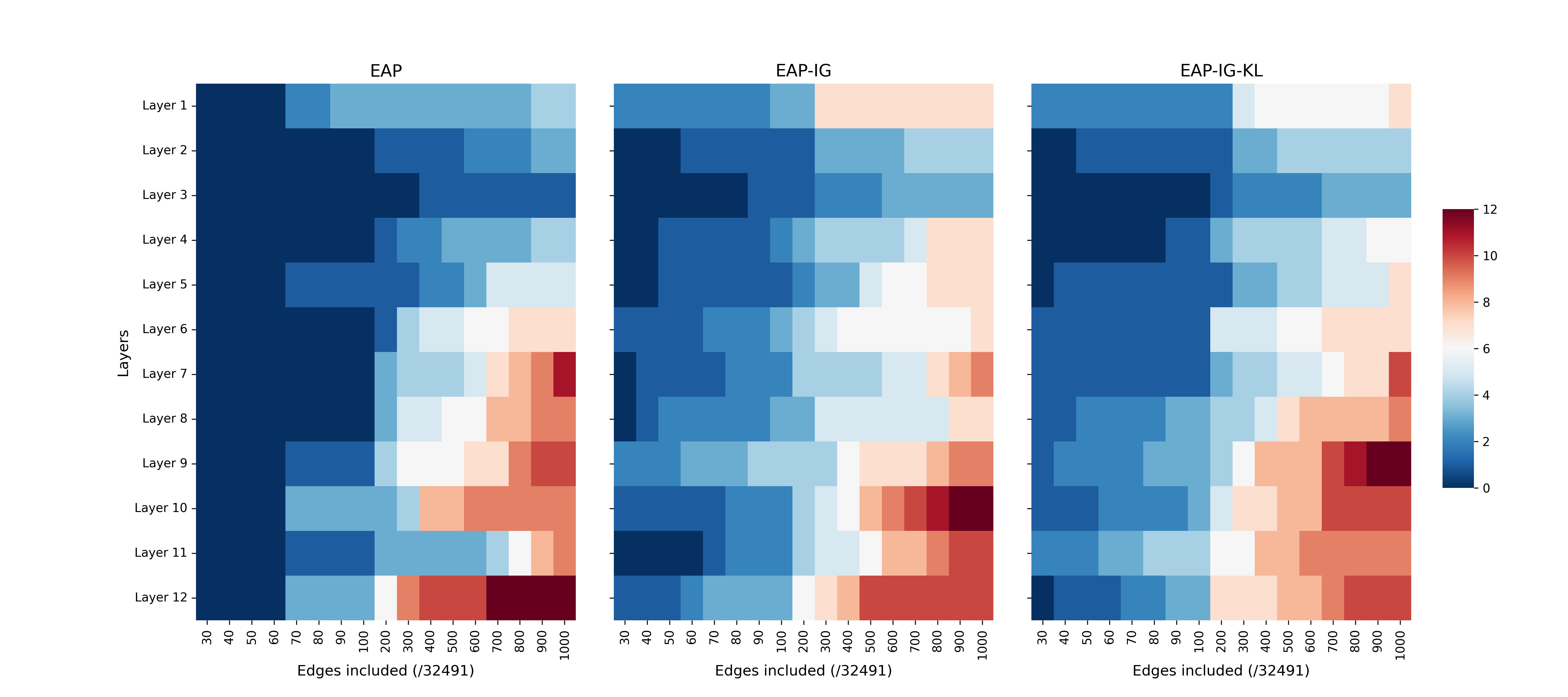}  
    \caption{Heatmap of node importance across layers for EAP, EAP-IG, and EAP-IG-KL methods. The x-axis shows the number of edges included, and the y-axis shows the layers. Darker colors represent higher node importance, with EAP focusing on the last layers and EAP-IG, EAP-IG-KL showing more balanced distributions across layers.}
    \label{fig:fig3}
\vspace{-12pt}
\end{figure*}

\paragraph{Dataset.}
We use the IOI dataset \citep{wang2022interpretability}, designed to evaluate models’ ability to perform indirect object identification tasks. Each entry consists of sentences with names and contexts, requiring the model to accurately predict the indirect object. The dataset contains minimal pairs of clean and corrupted inputs for direct comparison, testing the model’s robustness in distinguishing between potential candidates, even when distractor names are introduced. For more dataset and metric details, see Appendix 
 \ref{appendix:datasets}.

\paragraph{Baselines.} As SICAF is a mechanistic interpretation framework, we mainly implement it with previous Automatic Circuit-Finding approaches. Specifically, our baseline includes the following methods: EAP \citep{syed2023attribution}, which uses gradient-based approximations to estimate the importance of individual edges through their impact on model loss; EAP-IG \citep{hanna2024have}, which enhances EAP by employing integrated gradients along a path between clean and corrupted inputs to capture more accurate edge importance scores; and EAP-IG-KL \citep{hanna2024have}, which combines EAP-IG with Kullback-Leibler divergence as a loss function to generalize edge influence measurement across various interpretability tasks. Additional details are presented in Appendix \ref{appendix:Baselines}.

\noindent \textbf{Implementation Details.} We focus on simple tasks that are feasible even for GPT-2 small, which is the model most frequently studied from a circuits perspective. In our approach, we define GPT-2’s attention heads and MLPs (multi-layer perceptrons) as nodes within its computational graph. Additional implementation details can be found in Appendix \ref{appendix:Experimental Settings}.

\subsection{Circuit Identification and Faithfulness}

\begin{figure*}[ht]
    \centering   \includegraphics[width=0.8\linewidth]{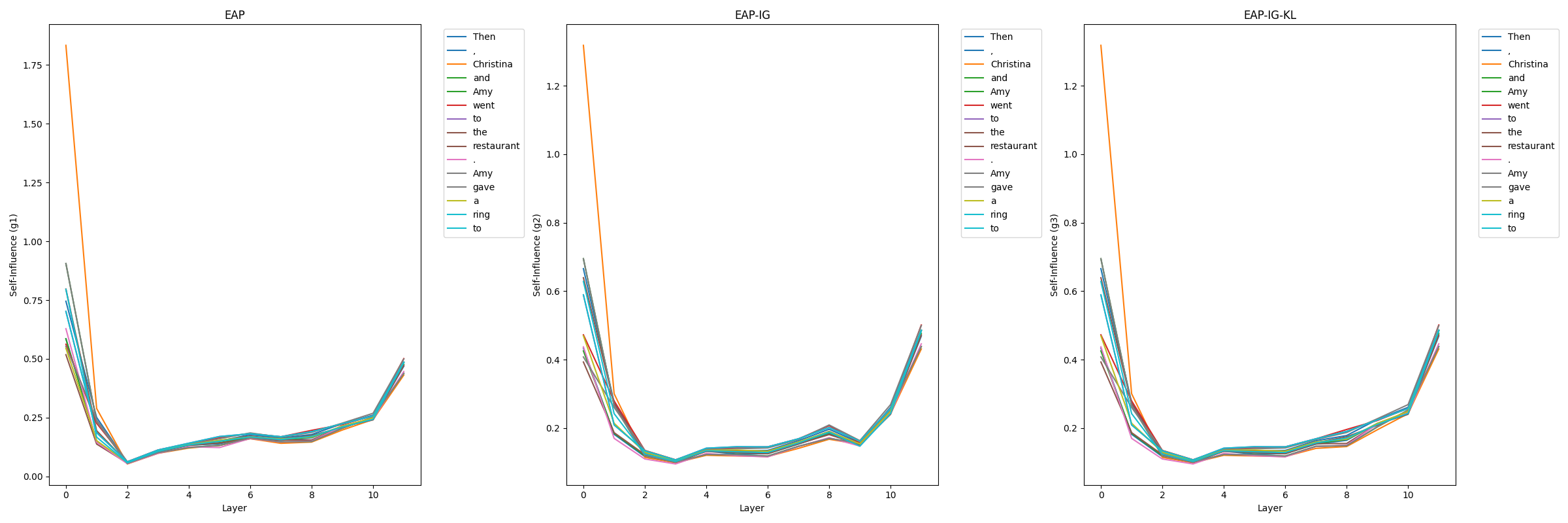}  
    \caption{Self-influence scores of key tokens across model layers for the EAP, EAP-IG, and EAP-IG-KL methods on the IOI task. Each subplot represents the distribution of self-influence for individual tokens across the 12 layers of the GPT-2 model. EAP shows concentrated influence in the early and final layers, while EAP-IG and EAP-IG-KL display more balanced self-influence across layers, reflecting a structured progression of token importance. Key tokens such as "Christina," "Amy," and "gave" consistently show high self-influence, demonstrating their significance in the reasoning process.}
    \label{fig:fig4}
\vspace{-5pt}
\end{figure*}

\begin{figure*}[ht]
    \centering
    \includegraphics[width=0.8\linewidth]{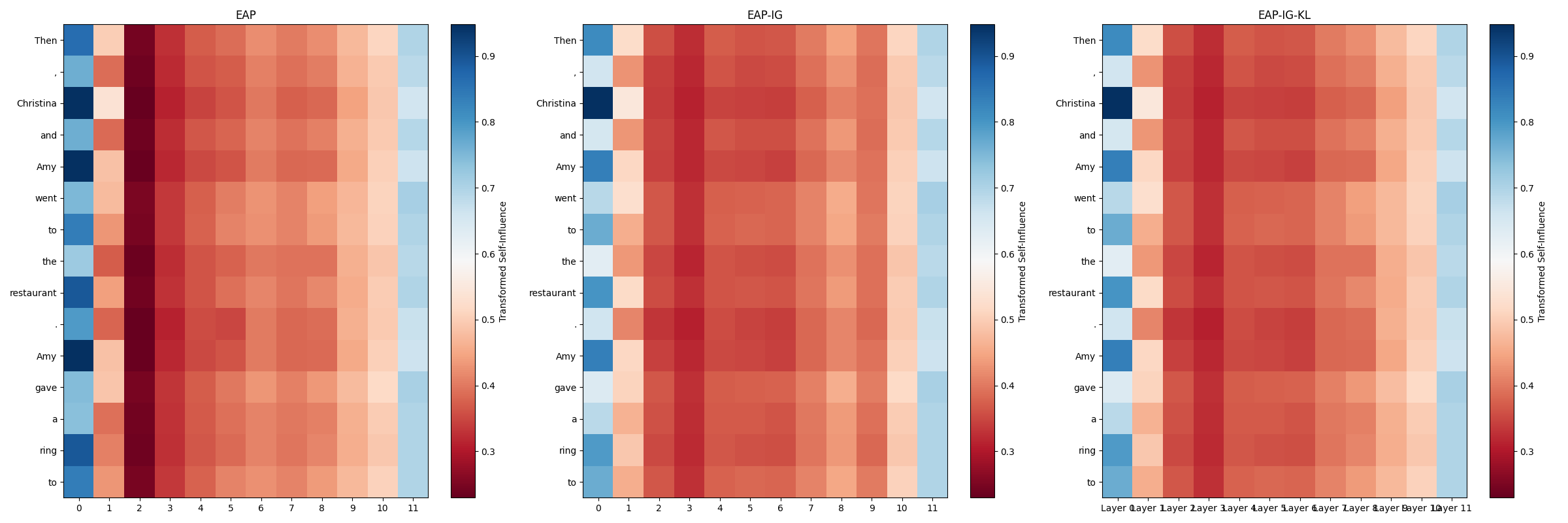}  
    \caption{Heatmap of node importance across layers for EAP, EAP-IG, and EAP-IG-KL methods. The x-axis represents the layer indices, and the y-axis shows the tokens. Darker colors indicate higher node importance, with EAP focusing on the last layers and EAP-IG and EAP-IG-KL exhibiting a more balanced distribution across layers.}
    \label{fig:fig5}
\vspace{-5pt}
\end{figure*}

Using the EAP, EAP-IG, and EAP-IG-KL methods, we successfully identified small circuits (containing 1-2\% of the edges) (see Table \ref{tab:100_edges_circuit} for the circuit composition at 100 edges for each method, with additional results in Appendix \ref{sec:appendixB}), recovering at least 85\% of the model’s performance on the IOI task. As shown in Figure \ref{fig:fig1} (left plot), both EAP-IG and EAP-IG-KL outperform EAP in terms of normalized faithfulness, achieving scores above 0.8 with approximately 800 edges, while EAP peaks below 0.6. This result suggests that EAP-IG and EAP-IG-KL are more effective at identifying circuits that retain model accuracy, particularly with fewer edges. Additionally, as seen in Figure \ref{fig:fig1} (center and right plots), EAP-IG and EAP-IG-KL exhibit rapid convergence in parameter inclusion, stabilizing around 35\% of the total parameters to achieve high performance, while EAP requires close to 40\% of the parameters to reach its maximum performance, which is still lower than the other methods. This result highlights the advantages of EAP-IG and EAP-IG-KL in identifying compact and faithful circuits, making them particularly suitable for tasks requiring high efficiency and accuracy.

\begin{table}[ht]
\centering
\caption{Circuit composition with 100 edges included using the greedy algorithm for EAP, EAP-IG, and EAP-IG-KL methods.}
\footnotesize
\renewcommand{\arraystretch}{1.2}
\begin{tabular}{l|p{0.3\textwidth}}
\toprule
\textbf{Method} & \textbf{Circuit Composition} \\
\midrule
EAP & input, a0.h1, a0.h10, m0, m4, a8.h10, a9.h3, a9.h4, m9, m10, a11.h2, a11.h3, m11, logits \\
EAP-IG & input, a0.h1, a0.h10, m0, m1, m2, a3.h0, m3, m4, a5.h5, a5.h9, m5, a6.h0, a6.h9, a7.h3, a7.h9, m7, a8.h3, a8.h6, a8.h10, m8, a9.h3, m9, a10.h2, m10, a11.h2, a11.h6, m11, logits \\
EAP-IG-KL & input, a0.h1, m0, m1, a3.h0, m4, a5.h5, a6.h9, a7.h1, a7.h3, a7.h9, a8.h6, a8.h10, m8, a9.h3, a9.h6, a9.h9, a10.h0, a10.h2, a10.h6, a10.h7, a11.h2, a11.h3, a11.h6, logits \\
\bottomrule
\end{tabular}
\label{tab:100_edges_circuit}
\end{table}

\subsection{Node Distribution Across Model Layers}
As shown in Figure \ref{fig:fig2}, the node distribution across model layers varies significantly between methods. EAP primarily activates nodes in the final layers (especially Layer 12), indicating that the model relies heavily on these layers to make final decisions in the IOI task. This pattern suggests that EAP emphasizes the semantic aggregation and decision generation occurring in the later layers. In contrast, EAP-IG and EAP-IG-KL demonstrate a more balanced node distribution across the early, middle, and final layers, with EAP-IG-KL showing substantial node activity between Layers 9 and 12. This pattern indicates that EAP-IG-KL captures more complex, multi-layered computational processes, utilizing information from a broader range of layers. Overall, the three methods share a consistent structure, with nodes concentrated in the first and last few layers, suggesting that these layers play a crucial role. The first layer likely helps in initial processing of input, while the last few layers appear critical for aggregating information before making final decisions.

\begin{table}[t]
\centering
\caption{Token influence scores for Layer 0 across EAP, EAP-IG, and EAP-IG-KL methods.}
\footnotesize
\renewcommand{\arraystretch}{1.2}
\begin{tabular}{l|c|c|c}
\toprule
\textbf{Token} & \textbf{EAP} & \textbf{EAP-IG} & \textbf{EAP-IG-KL} \\
\midrule
Then        & 0.745 & 0.666 & 0.666 \\
,           & 0.585 & 0.434 & 0.434 \\
Christina   & 1.834 & 1.318 & 1.318 \\
and         & 0.585 & 0.425 & 0.425 \\
Amy         & 0.906 & 0.695 & 0.695 \\
went        & 0.562 & 0.472 & 0.472 \\
to          & 0.702 & 0.589 & 0.589 \\
the         & 0.518 & 0.393 & 0.393 \\
restaurant  & 0.797 & 0.639 & 0.639 \\
.           & 0.628 & 0.437 & 0.437 \\
Amy         & 0.906 & 0.695 & 0.695 \\
gave        & 0.553 & 0.408 & 0.408 \\
\bottomrule
\end{tabular}
\label{tab:layer0_influence_scores}
\end{table}

\subsection{The Model's Thought Process}

Figures \ref{fig:fig3} and \ref{fig:fig4} provide insights into the layer-wise processing differences among the methods:EAP, EAP-IG, and EAP-IG-KL,specifically regarding their handling of self-influence scores. As shown in Figure \ref{fig:fig3}, EAP concentrates self-influence scores predominantly in the final layers, indicating a focus on quickly aggregating high-level information. This approach makes EAP suitable for tasks where immediate decision-making is crucial, as it enables rapid synthesis of key contextual elements. However, this reliance on the final layers limits EAP’s capability to capture intermediate reasoning steps, which are essential for handling more nuanced and multi-step reasoning tasks.

In contrast, Figure \ref{fig:fig4} illustrates that EAP-IG and EAP-IG-KL distribute influence scores more evenly across early, middle, and final layers, enabling a structured and gradual accumulation of information. This balanced distribution aligns well with tasks requiring multi-step reasoning, as it supports the retention and transformation of information throughout the model’s layered structure. EAP-IG-KL, in particular, achieves a high level of balance, ensuring stable self-influence scores across layers. This feature suggests that EAP-IG-KL is not only more robust in handling complex reasoning tasks but also better equipped to leverage information from both lower and higher layers.

The common hierarchical reasoning path followed by all three methods is best illustrated using the sentence “Then, Christina and Amy went to the restaurant. Amy gave a ring to...” as an example. Here, in the early layers, key entities like “Christina” and “Amy” are identified, setting a foundational context that informs subsequent reasoning steps. Moving to the middle layers, the model interprets the action verb “gave,” constructing relationships that frame “Amy” as the active entity in giving an item to another person. In the final layers, the model synthesizes this accumulated information, allowing it to conclude that “Amy” is the subject and infer the likely recipient.

The distinction among methods lies in the nuances of this shared reasoning path. EAP’s emphasis on final-layer influence results in faster decision-making but may overlook subtler contextual nuances essential for complex inferences. On the other hand, EAP-IG distributes influence with a greater emphasis on intermediate layers, focusing on refining relational structures and contextual relationships as the reasoning pathway progresses. EAP-IG-KL exhibits the most balanced distribution, making it highly adaptable for tasks that require intricate relationships and consistent reasoning across all layers.

The detailed self-influence scores in Table \ref{tab:layer0_influence_scores} further support this analysis. For instance, in Layer 0, the token “Christina” exhibits a high self-influence score, reinforcing its role as a key contextual marker that informs initial reasoning. As the model advances, tokens such as “gave” and “Amy” demonstrate increased influence in the final layers, highlighting their relevance in constructing the concluding inference. Additional token influence scores across layers and for other samples are available in Appendix \ref{appendix:Token self-Influence}, providing a comprehensive view of each method’s impact across the reasoning process.

\section{Conclusion}
We propose a new mechanistic interpretation framework, SICAF, to trace and analyze the thought processes that language models (LMs) employ during complex reasoning tasks, and we validate our approach on the GPT-2 model in the IOI reasoning task. By applying circuit analysis and self-influence functions, we successfully mapped the reasoning pathways within the GPT-2 model during the IOI task. Our method reveals a hierarchical structure in the model’s reasoning process, distinguishing key entities and relationships in a manner that resembles human reasoning steps. Overall, our findings contribute to a more interpretable and systematic understanding of the reasoning processes in language models, enhancing the transparency and trustworthiness of AI systems.

\section{Limitations}
While our study successfully elucidates certain thought processes within language models, it has some limitations. First, the analysis was conducted primarily on the GPT-2 model and may not generalize to larger or different architectures without adaptation. Additionally, calculating self-influence requires computationally intensive methods, which may pose scalability challenges for more complex models. Finally, our work focused on a single task (Indirect Object Identification, or IOI), and the applicability of these findings to other natural language processing tasks remains an open question. Future research should explore the adaptability of this approach across varied tasks and model architectures, as well as investigate methods to optimize computational efficiency.

\section*{Acknowledgement}
Di Wang and Lijie Hu are supported in part by the funding BAS/1/1689-01-01, URF/1/4663-01-01,  REI/1/5232-01-01,  REI/1/5332-01-01,  and URF/1/5508-01-01  from KAUST, and funding from KAUST - Center of Excellence for Generative AI, under award number 5940.

\bibliography{custom}
\appendix

\section{Additional Experiments}
\label{sec:appendix}
\subsection{Datasets}
\label{appendix:datasets}
The IOI (Indirect Object Identification) dataset (Wang et al., 2023) is designed to evaluate a model's ability to identify indirect objects within sentences. This task requires the model to recognize the indirect object among two specific names in a given sentence, predicting which name serves as the indirect object of the sentence. The dataset includes both clean and corrupted inputs. In clean inputs, each entry consists of a pair of names with contextual information, prompting the model to accurately identify the indirect object, or recipient, of an action. For example, in the sentence "When Amy and Laura got a snack at the house, Laura decided to give it to," the model should predict "Amy" as the indirect object. In corrupted inputs, the sentence structure remains unchanged, but the original indirect object is replaced by a third name, making both "Amy" and "Laura" approximately equally probable. For instance, replacing "Laura" with "Nicholas" yields "When Amy and Laura got a snack at the house, Nicholas decided to give it to," or replacing "Laura" with "Amy" results in "When Amy and Laura got a snack at the house, Amy decided to give it to," increasing the model's difficulty in distinguishing the correct indirect object. The model’s predictions are evaluated using logit difference (logit diff), calculated as the logit of the target indirect object minus the logit of the distractor. A larger logit difference indicates a stronger preference by the model for the correct indirect object. Using Wang et al.'s data generator, we generated a dataset of 1000 sentences table\ref{tab:ioi_dataset_examples}, containing both clean and corrupted inputs, for experimental analysis of the model's performance on this task.

\begin{table*}[b]
\footnotesize
\centering
\caption{Sample entries from the IOI (Indirect Object Identification) dataset, showcasing clean, corrupted, and corrupted hard inputs along with target and distractor indices.}
\renewcommand{\arraystretch}{1.2}
\begin{tabular}{>{\centering\arraybackslash}p{0.6\textwidth}|c|c}
\toprule
\multicolumn{3}{c}{\textbf{Clean Input}} \\
\midrule
\textbf{Sentence} & \textbf{Target Index} & \textbf{Distractor Index} \\
When Amy and Laura got a snack at the house, Laura decided to give it to & 14235 & 16753 \\
Then, Danielle and Andrew had a lot of fun at the office. Andrew gave a computer to & 39808 & 6858 \\
When Anthony and Jose got a drink at the restaurant, Jose decided to give it to & 9953 & 5264 \\
Then, Sean and Vanessa had a long argument, and afterwards Vanessa said to & 11465 & 42100 \\
\multicolumn{3}{c}{\dots \dots \dots \dots \dots \dots \dots \dots \dots \dots \dots \dots \dots \dots \dots} \\
\midrule
\multicolumn{3}{c}{\textbf{Corrupted Input}} \\
\midrule
\textbf{Sentence} & \textbf{Target Index} & \textbf{Distractor Index} \\
When Amy and Laura got a snack at the house, Nicholas decided to give it to & 14235 & 16753 \\
Then, Danielle and Andrew had a lot of fun at the office. Jeremy gave a computer to & 39808 & 6858 \\
When Anthony and Jose got a drink at the restaurant, Nathan decided to give it to & 9953 & 5264 \\
Then, Sean and Vanessa had a long argument, and afterwards Kimberly said to & 11465 & 42100 \\
\multicolumn{3}{c}{\dots \dots \dots \dots \dots \dots \dots \dots \dots \dots \dots \dots \dots \dots \dots} \\
\midrule
\multicolumn{3}{c}{\textbf{Corrupted Hard Input}} \\
\midrule
\textbf{Sentence} & \textbf{Target Index} & \textbf{Distractor Index} \\
When Amy and Laura got a snack at the house, Amy decided to give it to & 14235 & 16753 \\
Then, Danielle and Andrew had a lot of fun at the office. Danielle gave a computer to & 39808 & 6858 \\
When Anthony and Jose got a drink at the restaurant, Anthony decided to give it to & 9953 & 5264 \\
Then, Sean and Vanessa had a long argument, and afterwards Sean said to & 11465 & 42100 \\
\multicolumn{3}{c}{\dots \dots \dots \dots \dots \dots \dots \dots \dots \dots \dots \dots \dots \dots \dots} \\
\bottomrule
\end{tabular}
\label{tab:ioi_dataset_examples}
\end{table*}

\subsection{Details of Experimental Settings}
\label{appendix:Experimental Settings}
For the IOI task, we used a corpus of 1000 sentences created by \citep{hanna2024have}, based on the dataset generator from \citep{wang2022interpretability}, which includes both clean and corrupted inputs. We first fine-tuned GPT-2 on this dataset and then followed the experimental setup outlined in Michael Hanna's paper, defining GPT-2's attention heads and MLPs (multi-layer perceptrons) as nodes in its computational graph. A node's output directed to another node is defined as an edge. The number of nodes was 158, and the number of edges was 32,491. The input to node \(v\) is the sum of the outputs from all nodes \(u\) connected to \(v\). For \(n = 30, 40, \dots, 100, 200, \dots, 1000\), we selected a circuit of \(n\) edges using a greedy search procedure. Generally, larger \(n\) results in more faithful circuits, but these circuits tend to be less localized and interpretable. Let \(b\) and \(b'\) represent the model's performance on clean and corrupted inputs, respectively. The circuit's performance and faithfulness \(m\) are normalized as \((m - b') / (b - b')\).

\subsection{Baselines}
\label{appendix:Baselines}
\textbf{EAP}: EAP (Edge Attribution Patching) \citep{syed2023attribution}, which uses gradient-based approximations to assess the importance of individual edges by estimating the loss change from intervening on each model edge, allowing for a scalable approximation of causal effects without requiring extensive forward passes.

\noindent \textbf{EAP-IG}: EAP-IG (EAP with Integrated Gradients) \citep{hanna2024have}, which improves upon EAP by employing integrated gradients along a path between clean and corrupted inputs to compute a more faithful edge importance score, capturing a wider range of influence and avoiding zero-gradient issues common in standard gradient calculations.

\noindent \textbf{EAP-IG-KL}: EAP-IG-KL \citep{hanna2024have}, which combines EAP-IG with Kullback-Leibler (KL) divergence as a loss function, enabling it to generalize across tasks by measuring divergence between the patched and original model outputs, thus allowing consistent applicability to various interpretability tasks.

The table below (Table \ref{tab:eap_methods_comparison}) provides a comparison of the EAP, EAP-IG, and EAP-IG-KL methods, highlighting their distinctive attributes in terms of granularity, component, value, token positions, direction, and set used in the circuit-finding approach:

\begin{table*}[ht]
\footnotesize
\centering
\caption{Comparison of EAP, EAP-IG, and EAP-IG-KL methods. Each method differs in at least one aspect in terms of granularity, component, value, token positions, direction, and set.}
\renewcommand{\arraystretch}{0.8}
\begin{tabular}{l|p{1.8cm}|p{1.8cm}|p{2.6cm}|p{2.1cm}|p{2.1cm}|c}
\toprule
\textbf{Method} & \textbf{Granularity} & \textbf{Component} & \textbf{Value} & \textbf{Token Positions} & \textbf{Direction} & \textbf{Set} \\
\midrule
EAP        & Edges           & Edge      & Resample   & All tokens           & Resample Clean  & Circuit   \\
EAP-IG     & Edges           & Edge      & Integrated Gradient Path  & Specific tokens & Resample Clean & Circuit \\
EAP-IG-KL  & Edges           & Edge      & Integrated Gradient + KL Divergence & Specific tokens & Resample Clean & Circuit \\
\bottomrule
\end{tabular}
\label{tab:eap_methods_comparison}
\end{table*}

\section{Circuit Results}
\label{sec:appendixB}
\noindent "edges included": 200,\\
"greedy algorithm":\\
EAP: input, a0.h1, a0.h10, m0, m1, m3, m4, m5, a6.h0, a6.h9, m6, a7.h3, a7.h9, m7, a8.h3, a8.h6, a8.h10, m8, a9.h3, a9.h4, m9, a10.h2, a10.h7, m10, a11.h1, a11.h2, a11.h3, a11.h6, a11.h8, m11, logits\\
EAP-IG: input, a0.h1, a0.h10, m0, m1, m2, a3.h0, a3.h4, m3, a4.h3, m4, a5.h1, a5.h5, a5.h9, m5, a6.h0, a6.h6, a6.h9, m6, a7.h3, a7.h9, m7, a8.h3, a8.h6, a8.h10, m8, a9.h3, a9.h4, a9.h6, m9, a10.h0, a10.h2, a10.h7, m10, a11.h1, a11.h2, a11.h3, a11.h6, a11.h9, m11, logits\\
EAP-IG-KL: input, a0.h1, m0, m1, m2, a3.h0, a3.h4, m3, m4, a5.h1, a5.h5, a5.h8, a5.h9, m5, a6.h0, a6.h6, a6.h9, a7.h1, a7.h3, a7.h9, m7, a8.h3, a8.h6, a8.h10, m8, a9.h3, a9.h4, a9.h6, a9.h9, m9, a10.h0, a10.h1, a10.h2, a10.h6, a10.h7, m10, a11.h1, a11.h2, a11.h3, a11.h6, a11.h9, a11.h10, m11, logits\\

\vspace{1em}

\noindent "edges included": 300,\\
"greedy algorithm":\\
EAP: input, a0.h1, a0.h10, m0, m1, a3.h0, m3, m4, a5.h1, a5.h8, a5.h9, m5, a6.h0, a6.h6, a6.h9, m6, a7.h1, a7.h3, a7.h9, a7.h11, m7, a8.h3, a8.h5, a8.h6, a8.h10, a8.h11, m8, a9.h3, a9.h4, a9.h8, m9, a10.h2, a10.h7, m10, a11.h0, a11.h1, a11.h2, a11.h3, a11.h6, a11.h8, a11.h9, a11.h11, m11, logits\\
EAP-IG: input, a0.h1, a0.h3, a0.h4, a0.h5, a0.h9, a0.h10, m0, a1.h0, a1.h11, m1, a2.h2, m2, a3.h0, a3.h4, a3.h6, m3, a4.h3, a4.h4, m4, a5.h1, a5.h5, a5.h9, a5.h10, m5, a6.h0, a6.h6, a6.h9, m6, a7.h1, a7.h3, a7.h9, a7.h11, m7, a8.h3, a8.h6, a8.h10, m8, a9.h3, a9.h4, a9.h6, a9.h9, m9, a10.h0, a10.h2, a10.h6, a10.h7, m10, a11.h1, a11.h2, a11.h3, a11.h6, a11.h8, a11.h9, m11, logits\\
EAP-IG-KL: input, a0.h1, a0.h3, a0.h4, a0.h5, m0, a1.h0, a1.h11, m1, a2.h2, m2, a3.h0, a3.h4, a3.h6, m3, a4.h3, a4.h4, m4, a5.h1, a5.h5, a5.h8, a5.h9, m5, a6.h0, a6.h6, a6.h9, m6, a7.h1, a7.h3, a7.h7, a7.h9, m7, a8.h1, a8.h3, a8.h5, a8.h6, a8.h8, a8.h10, a8.h11, m8, a9.h2, a9.h3, a9.h4, a9.h6, a9.h7, a9.h9, m9, a10.h0, a10.h1, a10.h2, a10.h6, a10.h7, a10.h10, a10.h11, m10, a11.h1, a11.h2, a11.h3, a11.h6, a11.h9, a11.h10, m11, logits\\

\vspace{1em}

\noindent "edges included": 400,\\
"greedy algorithm":\\
EAP: input, a0.h1, a0.h10, m0, m1, m2, a3.h0, m3, a4.h4, m4, a5.h1, a5.h5, a5.h8, a5.h9, m5, a6.h0, a6.h6, a6.h9, m6, a7.h1, a7.h3, a7.h9, a7.h11, m7, a8.h3, a8.h5, a8.h6, a8.h10, a8.h11, m8, a9.h2, a9.h3, a9.h4, a9.h5, a9.h7, a9.h8, a9.h11, m9, a10.h2, a10.h7, m10, a11.h0, a11.h1, a11.h2, a11.h3, a11.h6, a11.h8, a11.h9, a11.h10, a11.h11, m11, logits\\
EAP-IG: input, a0.h1, a0.h3, a0.h4, a0.h5, a0.h9, a0.h10, m0, a1.h0, a1.h11, m1, a2.h2, m2, a3.h0, a3.h4, a3.h6, m3, a4.h3, a4.h4, m4, a5.h1, a5.h5, a5.h8, a5.h9, a5.h10, m5, a6.h0, a6.h6, a6.h9, m6, a7.h1, a7.h3, a7.h9, a7.h11, m7, a8.h3, a8.h5, a8.h6, a8.h10, a8.h11, m8, a9.h3, a9.h4, a9.h6, a9.h7, a9.h9, m9, a10.h0, a10.h2, a10.h6, a10.h7, m10, a11.h1, a11.h2, a11.h3, a11.h6, a11.h8, a11.h9, a11.h10, m11, logits\\
EAP-IG-KL: input, a0.h1, a0.h3, a0.h4, a0.h5, a0.h10, m0, a1.h0, a1.h11, m1, a2.h2, m2, a3.h0, a3.h4, a3.h6, m3, a4.h3, a4.h4, m4, a5.h1, a5.h5, a5.h8, a5.h9, m5, a6.h0, a6.h6, a6.h9, m6, a7.h1, a7.h3, a7.h6, a7.h7, a7.h9, a7.h11, m7, a8.h1, a8.h3, a8.h5, a8.h
\vspace{1em}

\noindent "edges included": 500,\\
"greedy algorithm":\\
EAP: input, a0.h1, a0.h10, m0, m1, m2, a3.h0, a3.h6, m3, a4.h4, m4, a5.h1, a5.h5, a5.h8, a5.h9, m5, a6.h0, a6.h6, a6.h9, m6, a7.h1, a7.h3, a7.h6, a7.h9, a7.h11, m7, a8.h3, a8.h5, a8.h6, a8.h10, a8.h11, m8, a9.h2, a9.h3, a9.h4, a9.h5, a9.h7, a9.h8, a9.h11, m9, a10.h2, a10.h7, m10, a11.h0, a11.h1, a11.h2, a11.h3, a11.h6, a11.h8, a11.h9, a11.h10, a11.h11, m11, logits\\
EAP-IG: input, a0.h1, a0.h3, a0.h4, a0.h5, a0.h9, a0.h10, m0, a1.h0, a1.h11, m1, a2.h2, m2, a3.h0, a3.h4, a3.h6, m3, a4.h3, a4.h4, a4.h6, a4.h7, m4, a5.h1, a5.h5, a5.h8, a5.h9, a5.h10, m5, a6.h0, a6.h6, a6.h9, m6, a7.h1, a7.h3, a7.h9, a7.h11, m7, a8.h1, a8.h3, a8.h5, a8.h6, a8.h10, a8.h11, m8, a9.h2, a9.h3, a9.h4, a9.h6, a9.h7, a9.h8, a9.h9, m9, a10.h0, a10.h2, a10.h6, a10.h7, a10.h10, m10, a11.h0, a11.h1, a11.h2, a11.h3, a11.h6, a11.h8, a11.h9, a11.h10, a11.h11, m11, logits\\
EAP-IG-KL: input, a0.h1, a0.h3, a0.h4, a0.h5, a0.h10, m0, a1.h0, a1.h5, a1.h11, m1, a2.h2, m2, a3.h0, a3.h4, a3.h6, m3, a4.h3, a4.h4, a4.h6, m4, a5.h1, a5.h5, a5.h8, a5.h9, a5.h10, m5, a6.h0, a6.h6, a6.h7, a6.h9, m6, a7.h1, a7.h3, a7.h6, a7.h7, a7.h9, a7.h11, m7, a8.h1, a8.h3, a8.h5, a8.h6, a8.h8, a8.h10, a8.h11, m8, a9.h0, a9.h2, a9.h3, a9.h4, a9.h6, a9.h7, a9.h9, m9, a10.h0, a10.h1, a10.h2, a10.h6, a10.h7, a10.h10, a10.h11, m10, a11.h1, a11.h2, a11.h3, a11.h6, a11.h8, a11.h9, a11.h10, m11, logits\\

\noindent \textit{(Results for edge values between 500 and 1000 follow similar patterns and are not listed in full here.)}

\vspace{1em}

\noindent "edges included": 1000,\\
"greedy algorithm":\\
EAP: input, a0.h1, a0.h9, a0.h10, m0, a1.h8, a1.h11, m1, m2, a3.h0, a3.h6, a3.h10, m3, a4.h3, a4.h4, a4.h6, a4.h7, m4, a5.h0, a5.h1, a5.h5, a5.h8, a5.h9, a5.h10, m5, a6.h0, a6.h1, a6.h2, a6.h3, a6.h4, a6.h5, a6.h6, a6.h8, a6.h9, a6.h10, m6, a7.h1, a7.h3, a7.h5, a7.h6, a7.h7, a7.h9, a7.h10, a7.h11, m7, a8.h0, a8.h1, a8.h3, a8.h5, a8.h6, a8.h7, a8.h9, a8.h10, a8.h11, m8, a9.h1, a9.h2, a9.h3, a9.h4, a9.h5, a9.h7, a9.h8, a9.h11, m9, a10.h0, a10.h1, a10.h2, a10.h3, a10.h4, a10.h7, a10.h8, a10.h10, m10, a11.h0, a11.h1, a11.h2, a11.h3, a11.h4, a11.h5, a11.h6, a11.h8, a11.h9, a11.h10, a11.h11, m11, logits\\
EAP-IG: input, a0.h1, a0.h3, a0.h4, a0.h5, a0.h9, a0.h10, m0, a1.h0, a1.h5, a1.h11, m1, a2.h2, a2.h9, m2, a3.h0, a3.h3, a3.h4, a3.h6, a3.h7, a3.h10, m3, a4.h3, a4.h4, a4.h6, a4.h7, a4.h8, a4.h11, m4, a5.h0, a5.h1, a5.h5, a5.h8, a5.h9, a5.h10, m5, a6.h0, a6.h1, a6.h4, a6.h5, a6.h6, a6.h7, a6.h8, a6.h9, m6, a7.h1, a7.h3, a7.h5, a7.h7, a7.h9
EAP-IG-KL: input, a0.h1, a0.h3, a0.h4, a0.h5, a0.h7, a0.h10, m0, a1.h0, a1.h5, a1.h11, m1, a2.h2, a2.h9, m2, a3.h0, a3.h3, a3.h4, a3.h6, a3.h7, m3, a4.h3, a4.h4, a4.h5, a4.h6, a4.h7, a4.h8, m4, a5.h1, a5.h5, a5.h8, a5.h9, a5.h10, a5.h11, m5, a6.h0, a6.h1, a6.h3, a6.h5, a6.h6, a6.h7, a6.h8, a6.h9, a6.h10, m6, a7.h1, a7.h3, a7.h5, a7.h6, a7.h7, a7.h8, a7.h9, a7.h11, m7, a8.h0, a8.h1, a8.h2, a8.h3, a8.h5, a8.h6, a8.h7, a8.h8, a8.h9, a8.h10, a8.h11, m8, a9.h0, a9.h2, a9.h3, a9.h4, a9.h5, a9.h6, a9.h7, a9.h9, a9.h11, m9, a10.h0, a10.h1, a10.h2, a10.h3, a10.h6, a10.h7, a10.h10, a10.h11, m10, a11.h0, a11.h1, a11.h2, a11.h3, a11.h6, a11.h8, a11.h9, a11.h10, a11.h11, m11, logits

\section{Token self-Influence}
\label{appendix:Token self-Influence}
See Tables \ref{tab1}, \ref{tab2}, and \ref{tab3} for complete results. Other results are similar and are not displayed.

\begin{table*}[ht]
\footnotesize
\caption{Self-influence scores across layers 0–11 using three different methods. The best influence score within each layer is highlighted in bold.}
\renewcommand{\arraystretch}{1.2}
\begin{center}
\resizebox{0.95\textwidth}{!}{ 
\begin{tabularx}{\textwidth}{p{0.11\textwidth}|p{0.05\textwidth}|p{0.05\textwidth}|p{0.05\textwidth}|p{0.05\textwidth}|p{0.05\textwidth}|p{0.05\textwidth}|p{0.05\textwidth}|p{0.05\textwidth}|p{0.05\textwidth}|p{0.05\textwidth}|p{0.05\textwidth}|p{0.05\textwidth}}
\toprule
\textbf{Token} & \textbf{L0} & \textbf{L1} & \textbf{L2} & \textbf{L3} & \textbf{L4} & \textbf{L5} & \textbf{L6} & \textbf{L7} & \textbf{L8} & \textbf{L9} & \textbf{L10} & \textbf{L11} \\
\midrule
\textbf{EAP} \\
Then & 0.745 & 0.251 & 0.059 & 0.107 & 0.138 & 0.150 & 0.178 & 0.163 & 0.178 & 0.224 & 0.261 & 0.484 \\
, & 0.585 & 0.150 & 0.056 & 0.103 & 0.132 & 0.139 & 0.166 & 0.153 & 0.164 & 0.214 & 0.246 & 0.469 \\
Christina & 1.834 & 0.289 & 0.052 & 0.098 & 0.120 & 0.132 & 0.160 & 0.141 & 0.146 & 0.197 & 0.242 & 0.431 \\
and & 0.585 & 0.148 & 0.057 & 0.104 & 0.133 & 0.143 & 0.168 & 0.155 & 0.167 & 0.213 & 0.246 & 0.476 \\
Amy & 0.906 & 0.233 & 0.055 & 0.101 & 0.123 & 0.132 & 0.163 & 0.147 & 0.149 & 0.205 & 0.252 & 0.438 \\
went & 0.562 & 0.226 & 0.062 & 0.112 & 0.139 & 0.165 & 0.183 & 0.168 & 0.196 & 0.221 & 0.258 & 0.501 \\
to & 0.702 & 0.186 & 0.061 & 0.112 & 0.141 & 0.170 & 0.181 & 0.169 & 0.190 & 0.224 & 0.257 & 0.486 \\
the & 0.518 & 0.138 & 0.056 & 0.104 & 0.132 & 0.142 & 0.161 & 0.154 & 0.155 & 0.212 & 0.240 & 0.472 \\
restaurant & 0.797 & 0.194 & 0.058 & 0.109 & 0.132 & 0.152 & 0.172 & 0.157 & 0.174 & 0.208 & 0.247 & 0.486 \\
. & 0.628 & 0.145 & 0.053 & 0.098 & 0.126 & 0.123 & 0.161 & 0.146 & 0.150 & 0.213 & 0.244 & 0.446 \\
Amy & 0.906 & 0.233 & 0.055 & 0.101 & 0.123 & 0.132 & 0.163 & 0.147 & 0.149 & 0.205 & 0.252 & 0.438 \\
gave & 0.553 & 0.240 & 0.062 & 0.110 & 0.137 & 0.160 & 0.185 & 0.167 & 0.187 & 0.227 & 0.268 & 0.499 \\
\midrule
\textbf{EAP-IG} \\
Then & 0.666 & 0.275 & 0.129 & 0.104 & 0.138 & 0.133 & 0.134 & 0.163 & 0.197 & 0.157 & 0.261 & 0.484 \\
, & 0.434 & 0.182 & 0.117 & 0.100 & 0.132 & 0.124 & 0.125 & 0.153 & 0.183 & 0.150 & 0.246 & 0.469 \\
Christina & 1.318 & 0.300 & 0.114 & 0.099 & 0.120 & 0.118 & 0.117 & 0.141 & 0.167 & 0.153 & 0.242 & 0.431 \\
and & 0.425 & 0.185 & 0.120 & 0.101 & 0.133 & 0.128 & 0.128 & 0.155 & 0.186 & 0.151 & 0.246 & 0.476 \\
Amy & 0.695 & 0.266 & 0.119 & 0.100 & 0.123 & 0.122 & 0.118 & 0.147 & 0.170 & 0.155 & 0.252 & 0.438 \\
went & 0.472 & 0.279 & 0.135 & 0.107 & 0.139 & 0.143 & 0.145 & 0.168 & 0.207 & 0.157 & 0.258 & 0.501 \\
to & 0.589 & 0.209 & 0.133 & 0.107 & 0.141 & 0.145 & 0.145 & 0.169 & 0.201 & 0.161 & 0.257 & 0.486 \\
the & 0.393 & 0.187 & 0.122 & 0.100 & 0.132 & 0.128 & 0.125 & 0.154 & 0.181 & 0.152 & 0.240 & 0.472 \\
restaurant & 0.639 & 0.271 & 0.125 & 0.106 & 0.132 & 0.134 & 0.133 & 0.157 & 0.190 & 0.152 & 0.247 & 0.486 \\
. & 0.437 & 0.170 & 0.110 & 0.095 & 0.126 & 0.119 & 0.116 & 0.146 & 0.172 & 0.147 & 0.244 & 0.446 \\
Amy & 0.695 & 0.266 & 0.119 & 0.100 & 0.123 & 0.122 & 0.118 & 0.147 & 0.170 & 0.155 & 0.252 & 0.438 \\
gave & 0.408 & 0.260 & 0.134 & 0.107 & 0.137 & 0.140 & 0.142 & 0.167 & 0.209 & 0.163 & 0.268 & 0.499 \\
\midrule
\textbf{EAP-IG-KL} \\
Then & 0.666 & 0.275 & 0.129 & 0.104 & 0.138 & 0.133 & 0.134 & 0.163 & 0.178 & 0.225 & 0.261 & 0.484 \\
, & 0.434 & 0.182 & 0.117 & 0.100 & 0.132 & 0.124 & 0.125 & 0.153 & 0.164 & 0.214 & 0.246 & 0.469 \\
Christina & 1.318 & 0.300 & 0.114 & 0.099 & 0.120 & 0.118 & 0.117 & 0.141 & 0.146 & 0.195 & 0.242 & 0.431 \\
and & 0.425 & 0.185 & 0.120 & 0.101 & 0.133 & 0.128 & 0.128 & 0.155 & 0.167 & 0.213 & 0.246 & 0.476 \\
Amy & 0.695 & 0.266 & 0.119 & 0.100 & 0.123 & 0.122 & 0.118 & 0.147 & 0.149 & 0.205 & 0.252 & 0.438 \\
went & 0.472 & 0.279 & 0.135 & 0.107 & 0.139 & 0.143 & 0.145 & 0.168 & 0.196 & 0.222 & 0.258 & 0.501 \\
to & 0.589 & 0.209 & 0.133 & 0.107 & 0.141 & 0.145 & 0.145 & 0.169 & 0.190 & 0.224 & 0.257 & 0.486 \\
the & 0.393 & 0.187 & 0.122 & 0.100 & 0.132 & 0.128 & 0.125 & 0.154 & 0.155 & 0.211 & 0.240 & 0.472 \\
restaurant & 0.639 & 0.271 & 0.125 & 0.106 & 0.132 & 0.134 & 0.133 & 0.157 & 0.174 & 0.208 & 0.247 & 0.486 \\
. & 0.437 & 0.170 & 0.110 & 0.095 & 0.126 & 0.119 & 0.116 & 0.146 & 0.150 & 0.213 & 0.244 & 0.446 \\
Amy & 0.695 & 0.266 & 0.119 & 0.100 & 0.123 & 0.122 & 0.118 & 0.147 & 0.149 & 0.205 & 0.252 & 0.438 \\
gave & 0.408 & 0.260 & 0.134 & 0.107 & 0.137 & 0.140 & 0.142 & 0.167 & 0.187 & 0.227 & 0.268 & 0.499 \\
\bottomrule
\end{tabularx}
}
\end{center}
\label{tab1}
\end{table*}

\begin{table*}[ht]
\footnotesize
\caption{Self-influence scores across layers 0–11 using three different methods. The best influence score within each layer is highlighted in bold.}
\renewcommand{\arraystretch}{}
\begin{center}
\resizebox{0.95\textwidth}{!}{ 
\begin{tabularx}{\textwidth}{p{0.11\textwidth}|p{0.05\textwidth}|p{0.05\textwidth}|p{0.05\textwidth}|p{0.05\textwidth}|p{0.05\textwidth}|p{0.05\textwidth}|p{0.05\textwidth}|p{0.05\textwidth}|p{0.05\textwidth}|p{0.05\textwidth}|p{0.05\textwidth}|p{0.05\textwidth}}
\toprule
\textbf{Token} & \textbf{L0} & \textbf{L1} & \textbf{L2} & \textbf{L3} & \textbf{L4} & \textbf{L5} & \textbf{L6} & \textbf{L7} & \textbf{L8} & \textbf{L9} & \textbf{L10} & \textbf{L11} \\
\midrule
\textbf{EAP}\\
Then & 0.797 & 0.533 & 0.058 & 0.102 & 1.571 & 0.130 & 0.171 & 0.168 & 0.174 & 0.214 & 0.337 & 0.463 \\
, & 0.615 & 0.501 & 0.055 & 0.098 & 1.507 & 0.123 & 0.161 & 0.160 & 0.162 & 0.200 & 0.316 & 0.449 \\
Danielle & 0.163 & 0.047 & 0.053 & 0.094 & 1.456 & 0.119 & 0.151 & 0.151 & 0.151 & 0.178 & 0.303 & 0.412 \\
and & 0.625 & 0.501 & 0.055 & 0.099 & 1.517 & 0.126 & 0.165 & 0.162 & 0.168 & 0.204 & 0.320 & 0.456 \\
Andrew & 0.618 & 0.480 & 0.053 & 0.097 & 1456 & 0.117 & 0.144 & 0.147 & 0.145 & 0.179 & 0.305 & 0.420 \\
had & 0.668 & 0.547 & 0.060 & 0.106 & 1.614 & 0.136 & 0.181 & 0.172 & 0.190 & 0.222 & 0.343 & 0.493 \\
a & 0.648 & 0.516 & 0.057 & 0.102 & 1.559 & 0.136 & 0.168 & 0.164 & 0.173 & 0.207 & 0.330 & 0.471 \\
lot & 0.803 & 0.533 & 0.059 & 0.104 & 1.594 & 0.134 & 0.176 & 0.171 & 0.184 & 0.217 & 0.341 & 0.484 \\
of & 0.667 & 0.510 & 0.056 & 0.099 & 1.522 & 0.127 & 0.159 & 0.161 & 0.161 & 0.201 & 0.323 & 0.457 \\
fun & 0.646 & 0.508 & 0.056 & 0.101 & 1.545 & 0.131 & 0.167 & 0.166 & 0.175 & 0.206 & 0.331 & 0.463 \\
at & 0.630 & 0.516 & 0.056 & 0.098 & 1.534 & 0.126 & 0.162 & 0.164 & 0.163 & 0.206 & 0.326 & 0.457 \\
the & 0.590 & 0.499 & 0.055 & 0.098 & 1.505 & 0.129 & 0.158 & 0.157 & 0.161 & 0.197 & 0.316 & 0.451 \\
office & 0.626 & 0.514 & 0.055 & 0.100 & 1.522 & 0.129 & 0.166 & 0.161 & 0.173 & 0.201 & 0.315 & 0.455 \\
. & 0.574 & 0.472 & 0.052 & 0.092 & 1.424 & 0.116 & 0.142 & 0.152 & 0.136 & 0.181 & 0.306 & 0.421 \\
Andrew & 0.618 & 0.489 & 0.053 & 0.097 & 1.456 & 0.117 & 0.144 & 0.147 & 0.145 & 0.179 & 0.305 & 0.420 \\
gave & 0.640 & 0.550 & 0.061 & 0.106 & 1.623 & 0.135 & 0.179 & 0.176 & 0.182 & 0.224 & 0.356 & 0.487 \\
a & 0.648 & 0.516 & 0.057 & 0.102 & 1.559 & 0.136 & 0.168 & 0.164 & 0.173 & 0.207 & 0.330 & 0.471 \\
computer & 0.711 & 0.519 & 0.058 & 0.106 & 1.586 & 0.136 & 0.174 & 0.168 & 0.179 & 0.210 & 0.337 & 0.482 \\
to & 0.909 & 0.570 & 0.064 & 0.113 & 1.714 & 0.143 & 0.192 & 0.186 & 0.189 & 0.235 & 0.340 & 0.460 \\

\midrule
\textbf{EAP-IG} \\
Then & 0.618 & 0.282 & 2.637 & 0.140 & 0.137 & 0.130 & 0.171 & 0.168 & 0.199 & 0.214 & 0.371 & 0.463 \\
, & 0.413 & 0.177 & 2.505 & 0.137 & 0.131 & 0.123 & 0.161 & 0.160 & 0.189 & 0.200 & 0.351 & 0.449 \\
Danielle & 1.090 & 0.327 & 2.399 & 0.125 & 0.123 & 0.119 & 0.151 & 0.151 & 0.180 & 0.178 & 0.339 & 0.412 \\
and & 0.415 & 0.172 & 2.510 & 0.138 & 0.134 & 0.126 & 0.165 & 0.162 & 0.193 & 0.204 & 0.356 & 0.456 \\
Andrew & 0.430 & 0.217 & 2.412 & 0.135 & 0.124 & 0.117 & 0.144 & 0.147 & 0.176 & 0.179 & 0.340 & 0.420 \\
had & 0.481 & 0.219 & 2.702 & 0.150 & 0.143 & 0.136 & 0.181 & 0.172 & 0.211 & 0.222 & 0.383 & 0.493 \\
a & 0.472 & 0.177 & 2.594 & 0.140 & 0.139 & 0.136 & 0.168 & 0.164 & 0.201 & 0.207 & 0.366 & 0.471 \\
lot & 0.478 & 0.240 & 2.655 & 0.143 & 0.141 & 0.134 & 0.176 & 0.171 & 0.207 & 0.217 & 0.376 & 0.484 \\
of & 0.438 & 0.169 & 2.526 & 0.144 & 0.129 & 0.127 & 0.159 & 0.161 & 0.190 & 0.201 & 0.354 & 0.457 \\
fun & 0.502 & 0.229 & 2.548 & 0.139 & 0.135 & 0.131 & 0.167 & 0.166 & 0.198 & 0.206 & 0.364 & 0.463 \\
at & 0.437 & 0.209 & 2.561 & 0.140 & 0.130 & 0.126 & 0.162 & 0.164 & 0.194 & 0.206 & 0.359 & 0.457 \\
the & 0.395 & 0.160 & 2.486 & 0.137 & 0.133 & 0.129 & 0.158 & 0.157 & 0.191 & 0.197 & 0.349 & 0.451 \\
office & 0.527 & 0.193 & 2.517 & 0.138 & 0.134 & 0.129 & 0.166 & 0.161 & 0.194 & 0.201 & 0.351 & 0.455 \\
. & 0.383 & 0.172 & 2.380 & 0.126 & 0.123 & 0.116 & 0.142 & 0.152 & 0.175 & 0.181 & 0.339 & 0.421 \\
Andrew & 0.430 & 0.217 & 2.412 & 0.135 & 0.124 & 0.117 & 0.144 & 0.147 & 0.176 & 0.179 & 0.340 & 0.420 \\
gave & 0.386 & 0.258 & 2.752 & 0.149 & 0.135 & 0.135 & 0.179 & 0.176 & 0.213 & 0.224 & 0.391 & 0.487 \\
a & 0.472 & 0.177 & 2.594 & 0.140 & 0.139 & 0.136 & 0.168 & 0.164 & 0.201 & 0.207 & 0.366 & 0.471 \\
computer & 0.545 & 0.214 & 2.635 & 0.149 & 0.139 & 0.136 & 0.174 & 0.168 & 0.202 & 0.210 & 0.373 & 0.482 \\
to & 0.657 & 0.223 & 2.860 & 0.151 & 0.152 & 0.143 & 0.192 & 0.186 & 0.217 & 0.235 & 0.379 & 0.460 \\
\midrule
\textbf{EAP-IG-KL} \\
Then & 0.617 & 0.290 & 2.637 & 0.140 & 0.137 & 0.130 & 0.171 & 0.168 & 0.199 & 0.222 & 0.219 & 0.560 \\
, & 0.413 & 0.189 & 2.505 & 0.137 & 0.131 & 0.123 & 0.161 & 0.160 & 0.189 & 0.210 & 0.208 & 0.526 \\
Danielle & 1.090 & 0.339 & 2.399 & 0.125 & 0.123 & 0.119 & 0.151 & 0.151 & 0.180 & 0.199 & 0.213 & 0.509 \\
and & 0.415 & 0.191 & 2.510 & 0.138 & 0.134 & 0.126 & 0.165 & 0.162 & 0.193 & 0.209 & 0.209 & 0.531 \\
Andrew & 0.430 & 0.240 & 2.412 & 0.135 & 0.124 & 0.117 & 0.144 & 0.147 & 0.176 & 0.213 & 0.213 & 0.529 \\
had & 0.481 & 0.259 & 2.702 & 0.150 & 0.143 & 0.136 & 0.181 & 0.172 & 0.211 & 0.217 & 0.220 & 0.558 \\
a & 0.472 & 0.212 & 2.594 & 0.140 & 0.139 & 0.136 & 0.168 & 0.164 & 0.201 & 0.209 & 0.210 & 0.539 \\
lot & 0.478 & 0.269 & 2.655 & 0.143 & 0.141 & 0.134 & 0.176 & 0.171 & 0.207 & 0.217 & 0.215 & 0.540 \\
of & 0.438 & 0.186 & 2.526 & 0.144 & 0.129 & 0.127 & 0.159 & 0.161 & 0.190 & 0.211 & 0.208 & 0.537 \\
fun & 0.502 & 0.258 & 2.548 & 0.139 & 0.135 & 0.131 & 0.167 & 0.166 & 0.198 & 0.205 & 0.207 & 0.529 \\
at & 0.437 & 0.220 & 2.561 & 0.140 & 0.130 & 0.126 & 0.162 & 0.164 & 0.194 & 0.217 & 0.214 & 0.537 \\
the & 0.395 & 0.189 & 2.486 & 0.137 & 0.133 & 0.129 & 0.158 & 0.157 & 0.191 & 0.208 & 0.204 & 0.526 \\
office & 0.527 & 0.233 & 2.517 & 0.138 & 0.134 & 0.129 & 0.166 & 0.161 & 0.194 & 0.201 & 0.201 & 0.509 \\
. & 0.383 & 0.173 & 2.380 & 0.126 & 0.123 & 0.116 & 0.142 & 0.152 & 0.175 & 0.207 & 0.205 & 0.514 \\
Andrew & 0.430 & 0.240 & 2.412 & 0.135 & 0.124 & 0.117 & 0.144 & 0.147 & 0.176 & 0.213 & 0.213 & 0.529 \\
gave & 0.386 & 0.280 & 2.752 & 0.149 & 0.135 & 0.135 & 0.179 & 0.176 & 0.213 & 0.227 & 0.230 & 0.577 \\
a & 0.472 & 0.212 & 2.594 & 0.140 & 0.139 & 0.136 & 0.168 & 0.164 & 0.201 & 0.209 & 0.210 & 0.539 \\
computer & 0.545 & 0.255 & 2.635 & 0.149 & 0.139 & 0.136 & 0.174 & 0.168 & 0.202 & 0.216 & 0.214 & 0.546 \\
to & 0.657 & 0.211 & 2.860 & 0.151 & 0.152 & 0.143 & 0.192 & 0.186 & 0.217 & 0.227 & 0.229 & 0.533 \\
\bottomrule
\end{tabularx}
}
\end{center}
\label{tab2}
\end{table*}

\begin{table*}[ht]
\footnotesize
\caption{Self-influence scores across layers 0–11 using EAP method. The best influence score within each layer is highlighted.}
\renewcommand{\arraystretch}{1.2}
\begin{center}
\resizebox{0.95\textwidth}{!}{ 
\begin{tabularx}{\textwidth}{p{0.11\textwidth}|p{0.05\textwidth}|p{0.05\textwidth}|p{0.05\textwidth}|p{0.05\textwidth}|p{0.05\textwidth}|p{0.05\textwidth}|p{0.05\textwidth}|p{0.05\textwidth}|p{0.05\textwidth}|p{0.05\textwidth}|p{0.05\textwidth}|p{0.05\textwidth}}
\toprule
\textbf{Token} & \textbf{L0} & \textbf{L1} & \textbf{L2} & \textbf{L3} & \textbf{L4} & \textbf{L5} & \textbf{L6} & \textbf{L7} & \textbf{L8} & \textbf{L9} & \textbf{L10} & \textbf{L11} \\
\midrule
\textbf{EAP} \\
\text{When} & 0.533 & 0.514 & 0.056 & 0.093 & 1.478 & 0.116 & 0.172 & 0.153 & 0.173 & 0.215 & 0.346 & 0.461 \\
\text{Anthony} & 1.255 & 0.441 & 0.049 & 0.083 & 1.331 & 0.104 & 0.149 & 0.132 & 0.148 & 0.181 & 0.310 & 0.395 \\
\text{and} & 0.390 & 0.468 & 0.053 & 0.089 & 1.430 & 0.112 & 0.167 & 0.145 & 0.171 & 0.207 & 0.322 & 0.448 \\
\text{Jose} & 0.841 & 0.439 & 0.050 & 0.084 & 1.353 & 0.107 & 0.156 & 0.138 & 0.154 & 0.188 & 0.315 & 0.407 \\
\text{got} & 0.533 & 0.513 & 0.057 & 0.096 & 1.500 & 0.120 & 0.179 & 0.158 & 0.184 & 0.222 & 0.342 & 0.473 \\
\text{a} & 0.601 & 0.490 & 0.055 & 0.093 & 1.480 & 0.119 & 0.173 & 0.147 & 0.177 & 0.210 & 0.335 & 0.470 \\
\text{drink} & 0.537 & 0.485 & 0.055 & 0.095 & 1.490 & 0.120 & 0.176 & 0.155 & 0.182 & 0.215 & 0.332 & 0.470 \\
\text{at} & 0.489 & 0.479 & 0.054 & 0.088 & 1.434 & 0.110 & 0.159 & 0.144 & 0.166 & 0.206 & 0.327 & 0.445 \\
\text{the} & 0.486 & 0.469 & 0.052 & 0.088 & 1.417 & 0.111 & 0.163 & 0.140 & 0.164 & 0.199 & 0.319 & 0.448 \\
\text{restaurant} & 0.712 & 0.481 & 0.055 & 0.095 & 1.485 & 0.121 & 0.176 & 0.153 & 0.180 & 0.212 & 0.326 & 0.459 \\
, & 0.423 & 0.465 & 0.052 & 0.088 & 1.408 & 0.110 & 0.161 & 0.142 & 0.165 & 0.203 & 0.319 & 0.436 \\
\text{Jose} & 0.841 & 0.439 & 0.050 & 0.084 & 1.353 & 0.107 & 0.156 & 0.138 & 0.154 & 0.188 & 0.315 & 0.407 \\
\text{decided} & 0.816 & 0.534 & 0.061 & 0.101 & 1.601 & 0.129 & 0.187 & 0.167 & 0.197 & 0.236 & 0.365 & 0.499 \\
\text{to} & 0.658 & 0.533 & 0.060 & 0.096 & 1.605 & 0.130 & 0.182 & 0.158 & 0.194 & 0.221 & 0.352 & 0.481 \\
\text{give} & 0.421 & 0.523 & 0.058 & 0.097 & 1.528 & 0.120 & 0.180 & 0.159 & 0.188 & 0.228 & 0.355 & 0.484 \\
\text{it} & 0.653 & 0.516 & 0.057 & 0.101 & 1.544 & 0.131 & 0.188 & 0.161 & 0.196 & 0.227 & 0.342 & 0.484 \\
\text{to} & 0.658 & 0.533 & 0.060 & 0.096 & 1.605 & 0.130 & 0.182 & 0.158 & 0.194 & 0.221 & 0.352 & 0.481 \\
\midrule
\textbf{EAP-IG} \\
\text{When} & 0.391 & 0.278 & 2.545 & 0.140 & 0.111 & 0.116 & 0.172 & 0.153 & 0.186 & 0.215 & 0.377 & 0.461 \\
\text{Anthony} & 0.878 & 0.269 & 2.223 & 0.128 & 0.099 & 0.104 & 0.149 & 0.132 & 0.154 & 0.181 & 0.331 & 0.395 \\
\text{and} & 0.321 & 0.175 & 2.388 & 0.129 & 0.108 & 0.112 & 0.167 & 0.145 & 0.175 & 0.207 & 0.355 & 0.448 \\
\text{Jose} & 0.643 & 0.264 & 2.256 & 0.127 & 0.102 & 0.107 & 0.156 & 0.138 & 0.159 & 0.188 & 0.341 & 0.407 \\
\text{got} & 0.450 & 0.243 & 2.559 & 0.140 & 0.114 & 0.120 & 0.179 & 0.158 & 0.190 & 0.222 & 0.377 & 0.473 \\
\text{a} & 0.440 & 0.191 & 2.488 & 0.136 & 0.114 & 0.119 & 0.173 & 0.147 & 0.183 & 0.210 & 0.369 & 0.470 \\
\text{drink} & 0.554 & 0.231 & 2.491 & 0.136 & 0.114 & 0.120 & 0.176 & 0.155 & 0.187 & 0.215 & 0.365 & 0.470 \\
\text{at} & 0.397 & 0.222 & 2.425 & 0.131 & 0.106 & 0.110 & 0.159 & 0.144 & 0.176 & 0.206 & 0.358 & 0.445 \\
\text{the} & 0.365 & 0.171 & 2.370 & 0.132 & 0.106 & 0.111 & 0.163 & 0.140 & 0.170 & 0.199 & 0.351 & 0.448 \\
\text{restaurant} & 0.569 & 0.233 & 2.469 & 0.136 & 0.112 & 0.121 & 0.176 & 0.153 & 0.183 & 0.212 & 0.361 & 0.459 \\
, & 0.336 & 0.180 & 2.364 & 0.129 & 0.105 & 0.110 & 0.161 & 0.142 & 0.173 & 0.203 & 0.348 & 0.436 \\
\text{Jose} & 0.643 & 0.264 & 2.256 & 0.127 & 0.102 & 0.107 & 0.156 & 0.138 & 0.159 & 0.188 & 0.341 & 0.407 \\
\text{decided} & 0.720 & 0.270 & 2.722 & 0.145 & 0.122 & 0.129 & 0.187 & 0.167 & 0.205 & 0.236 & 0.399 & 0.499 \\
\text{to} & 0.566 & 0.208 & 2.681 & 0.134 & 0.125 & 0.130 & 0.182 & 0.158 & 0.195 & 0.221 & 0.376 & 0.481 \\
\text{give} & 0.320 & 0.259 & 2.622 & 0.142 & 0.115 & 0.120 & 0.180 & 0.159 & 0.196 & 0.228 & 0.387 & 0.484 \\
\text{it} & 0.562 & 0.209 & 2.580 & 0.138 & 0.123 & 0.131 & 0.188 & 0.161 & 0.197 & 0.227 & 0.375 & 0.484 \\
\text{to} & 0.566 & 0.208 & 2.681 & 0.134 & 0.125 & 0.130 & 0.182 & 0.158 & 0.195 & 0.221 & 0.376 & 0.481 \\
\midrule
\textbf{EAP-IG-KL} \\
\text{When} & 0.391 & 0.256 & 2.545 & 0.140 & 0.111 & 0.116 & 0.172 & 0.153 & 0.186 & 0.192 & 0.258 & 0.581 \\
\text{Anthony} & 0.878 & 0.256 & 2.223 & 0.128 & 0.099 & 0.104 & 0.149 & 0.132 & 0.154 & 0.173 & 0.229 & 0.510 \\
\text{and} & 0.321 & 0.180 & 2.388 & 0.129 & 0.108 & 0.112 & 0.167 & 0.145 & 0.175 & 0.173 & 0.236 & 0.530 \\
\text{Jose} & 0.643 & 0.251 & 2.256 & 0.127 & 0.102 & 0.107 & 0.156 & 0.138 & 0.159 & 0.172 & 0.234 & 0.514 \\
\text{got} & 0.450 & 0.247 & 2.559 & 0.140 & 0.114 & 0.120 & 0.179 & 0.158 & 0.190 & 0.185 & 0.248 & 0.561 \\
\text{a} & 0.440 & 0.224 & 2.488 & 0.136 & 0.114 & 0.119 & 0.173 & 0.147 & 0.183 & 0.172 & 0.247 & 0.542 \\
\text{drink} & 0.554 & 0.248 & 2.491 & 0.136 & 0.114 & 0.120 & 0.176 & 0.155 & 0.187 & 0.176 & 0.235 & 0.536 \\
\text{at} & 0.397 & 0.212 & 2.425 & 0.131 & 0.106 & 0.110 & 0.159 & 0.144 & 0.176 & 0.175 & 0.242 & 0.544 \\
\text{the} & 0.365 & 0.192 & 2.370 & 0.132 & 0.106 & 0.111 & 0.163 & 0.140 & 0.170 & 0.169 & 0.236 & 0.526 \\
\text{restaurant} & 0.569 & 0.258 & 2.469 & 0.136 & 0.112 & 0.121 & 0.176 & 0.153 & 0.183 & 0.174 & 0.234 & 0.527 \\
, & 0.336 & 0.177 & 2.364 & 0.129 & 0.105 & 0.110 & 0.161 & 0.142 & 0.173 & 0.174 & 0.236 & 0.526 \\
\text{Jose} & 0.643 & 0.251 & 2.256 & 0.127 & 0.102 & 0.107 & 0.156 & 0.138 & 0.159 & 0.172 & 0.234 & 0.514 \\
\text{decided} & 0.720 & 0.275 & 2.722 & 0.145 & 0.122 & 0.129 & 0.187 & 0.167 & 0.205 & 0.196 & 0.260 & 0.581 \\
\text{to} & 0.566 & 0.215 & 2.681 & 0.134 & 0.125 & 0.130 & 0.182 & 0.158 & 0.195 & 0.185 & 0.258 & 0.535 \\
\text{give} & 0.320 & 0.253 & 2.622 & 0.142 & 0.115 & 0.120 & 0.180 & 0.159 & 0.196 & 0.192 & 0.254 & 0.579 \\
\text{it} & 0.562 & 0.240 & 2.580 & 0.138 & 0.123 & 0.131 & 0.188 & 0.161 & 0.197 & 0.182 & 0.241 & 0.544 \\
\text{to} & 0.566 & 0.215 & 2.681 & 0.134 & 0.125 & 0.130 & 0.182 & 0.158 & 0.195 & 0.185 & 0.258 & 0.535 \\
\bottomrule
\end{tabularx}
}
\end{center}
\label{tab3}
\end{table*}

\begin{algorithm}[b]
\caption{Self-Influence Circuit Analysis Framework}
\label{alg:alg1}
\KwIn{Model $M$, Input sample $x$, Circuit-finding methods $\{ \text{EAP}, \text{EAP-IG}, \text{EAP-IG-KL} \}$, Scaling factor $c$}
\KwOut{Inferred thought process structure}

\textbf{Initialize} model parameters $\theta$\;
\textbf{Select} circuit-finding method (e.g., EAP, EAP-IG, or EAP-IG-KL)\;

\textbf{Phase 1: Automatic Circuit Identification} \\
\SetKwFunction{FIdent}{IdentifyCriticalCircuits}
\SetKwProg{Fn}{Function}{:}{}
\Fn{\FIdent{$M$, $x$, method}}{
    \For{each edge $(u, v)$ in $M$}{
        Perturb edge $(u, v)$ to evaluate effect on loss $L(x)$\;
        Calculate edge importance score $S_{uv}$ based on perturbation: \[
        S_{uv} = \Delta L(x; u, v)
        \]
    }
    Rank edges by importance and select top-$k$\;
    \KwRet CircuitGraph (subgraph with highest-ranked edges)\;
}

CircuitGraph $\leftarrow$ \FIdent{$M$, $x$, chosen method}\;

\textbf{Phase 2: Layer-wise Self-Influence Computation} \\
\SetKwFunction{FComp}{ComputeSelfInfluence}
\Fn{\FComp{CircuitGraph, $x$, $\theta$, $c$}}{
    \For{each layer $\ell$ in CircuitGraph}{
        Initialize self-influence $I_\ell(x, x) \leftarrow 0$\;
        Compute gradient $\nabla_{\theta} L(x)$ w.r.t. parameters $\theta$ at layer $\ell$\;
        Calculate Hessian matrix $H$ based on $\nabla_{\theta} L(x)$\;
        \If{$||H|| > 1$}{
            Scale Hessian by factor $c$: $H \leftarrow cH$\;
        }
        Approximate $H^{-1}$ using Taylor expansion: \[
        H^{-1} \approx \sum_{i=0}^{\infty} (I - H)^i
        \]
        Calculate self-influence for layer $\ell$: \[
        I_\ell(x, x) = - \nabla_{\theta} L(x)^\top H^{-1} \nabla_{\theta} L(x)
        \]
    }
    \KwRet LayerwiseInfluenceScores for all layers\;
}

LayerwiseInfluenceScores $\leftarrow$ \FComp{CircuitGraph, $x$, $\theta$, $c$}\;

\textbf{Phase 3: Thought Process Inference} \\
\SetKwFunction{FInfer}{InferThoughtProcess}
\Fn{\FInfer{LayerwiseInfluenceScores}}{
    Initialize ThoughtProcess $\leftarrow \{\}$\;
    \For{each layer $\ell$}{
        Analyze distribution $\{ I_\ell(x, x_i) \}_{i=1}^n$ across tokens\;
        Identify significant contributions to model’s decision pathway at layer $\ell$\;
        Update ThoughtProcess with token importance at each layer\;
    }
    \KwRet ThoughtProcess\;
}

ThoughtProcess $\leftarrow$ \FInfer{LayerwiseInfluenceScores}\;

\KwRet ThoughtProcess\;
\end{algorithm}

\end{document}